\documentclass{article}

\usepackage{arxiv}

\usepackage[utf8]{inputenc} 
\usepackage[T1]{fontenc}    
\usepackage{hyperref}       
\usepackage{url}            
\usepackage{booktabs}       
\usepackage{amsfonts}       
\usepackage{nicefrac}       
\usepackage{microtype}      
\usepackage{lipsum}
\usepackage{graphicx}
\graphicspath{ {./images/} }
\usepackage{mathtools}
\usepackage{multirow} 
\usepackage{natbib}
\usepackage{algorithm}
\usepackage{algpseudocode}
\usepackage{graphicx}
\graphicspath{ {./images/} }
\usepackage{listings} 

\title{Reconstructing a dynamical system and forecasting time series by self-consistent deep learning}

\author{ {Zhe Wang}\\
	Energy research institute @NTU\\
	Nanyang Technological University\\
	637141 Singapore \\
	\texttt{zhe.wang@ntu.edu.sg} \\
	\And
	{Claude Guet} \\
	Energy research institute @NTU\\
	Nanyang Technological University\\
	637141 Singapore \\
	and School of Materials Science and Engineering\\
	Nanyang Technological University\\ 
	639798 Singapore\\
	\texttt{cguet@ntu.edu.sg} \\}

\renewcommand{\shorttitle}{Deep dynamical system reconstruction \& forecasting}

\begin{document}
\maketitle

\begin{abstract}
	We introduce a self-consistent deep-learning framework which, for a noisy deterministic time series, provides unsupervised filtering, state space reconstruction, identification of the underlying differential equations and forecasting. Without a priori information on the signal, we embed the time series in a state space, where deterministic structures, i.e. attractors, are revealed. Under the assumption that the evolution of solution trajectories is described by an unknown dynamical system, we filter out stochastic outliers. The embedding function, the solution trajectories and the dynamical systems are constructed using deep neural networks, respectively. By exploiting the differentiability of the neural solution trajectory, the neural dynamical system is defined locally at each time, mitigating the need for propagating gradients through numerical solvers. On a chaotic time series masked by additive Gaussian noise, we demonstrate the filtering ability and the predictive power of the proposed framework.
\end{abstract}


\section{Introduction}
Time series analysis and time series forecasting have been studied to extract information about the data and the underlying dynamics, and to predict the future of observables from past measurements. The first systematic modeling of time series dates back to 1927 when \citet{Yule27} introduced  a linear autoregression model to reveal the dynamics of sunspot numbers. The model which writes as
\begin{align} \label{eqn AR}
    u(k+1) = \sum_{n=0}^{m-1} \boldsymbol{a}(n) u(k-n) + \boldsymbol{e}(k),
\end{align}
takes the form of a linear difference equation, which states that the future $u(k+1)$ is a weighted sum of the past values in the sequence, with the error term $\boldsymbol{e}(k)$. Here, $m$ denotes the regression order.

Since linear equations lead to only exponential or periodic motions, analyzing systems characterized by irregular motions calls for nonlinear autoregression models which advantageously employ neural networks. Retaining the basic form of Eq. (\ref{eqn AR}), neural autoregression models can be expressed as
\begin{align} \label{eqn NAR}
    u(k+1) = \mathcal{N}[u(k), u(k-1), ..., u(k-(m-1)); \boldsymbol{\theta}] + \boldsymbol{e}(k),
\end{align}
where the model parameters $\boldsymbol{\theta}$ of the neural network, denoted by $\mathcal{N}(\cdot)$, are  determined by minimizing the deviation $\boldsymbol{e}(k)$. Since the first application of neural networks to reconstruct the governing equations underlying a chaotic time series by \citet{Lapedes87}, various network architectures including multilayer perceptrons \citep{Weigend90,Weigend91}, time delayed neural networks \citep{Wan94,Wunsch98}, convolution neural networks \citep{Borovykh17}, recurrent neural networks \citep{Gers02,LeCun09,Dubois20}, and more recently transformers \citep{Li19,Lim2021}, have been explored to model time series arising from physics, engineering, biological science, finance,...

Using neural networks as a nonlinear autoregression model stems from the universal approximation theorem which states that a sufficiently deep neural network can approximate any arbitrary well-behaved nonlinear function with a finite set of parameters \citep{Gorban98,Winkler17,Lin18} and from the \citet{Takens81} theorem in the following manner. Let $\boldsymbol{v}(t)$ be a state vector on the solution manifold and let 
\begin{align}
    \frac{\mathrm{d} \boldsymbol{v}(t)}{\mathrm{d} t} = f[\boldsymbol{v}(t)],
\end{align}
be the governing equation. One seldom has access to $\boldsymbol{v}(t)$. Instead, the state of dynamical systems is partially observed and inferred through a sequence of scalar measurements $u(k)$ sampled at discrete times and, frequently, masked by noise. For a sufficiently large dimension $m \in \mathbb{Z}^+$ and an arbitrary delay time $\tau \in \mathbb{R}^+$, \citet{Takens81} theorem affirms the existence of a diffeomorphism between a delay vector $\tilde{\boldsymbol{u}}$ and the underlying state $\boldsymbol{v}(t)$ of the dynamical system. This implies that there exists a nonlinear mapping $u(k+1) = g[\tilde{\boldsymbol{u}}(k)]$ which models the time series exactly. In virtue of the universal approximation theorem, neural networks could capture the $g(\cdot)$ mapping.

Going beyond merely finding the underlying difference equations (\ref{eqn NAR}), inspired by recent works in learning differential equations \citep{Chen2018, Ayed19, Rassi19, Rackauckas20}, we propose to reconstruct a latent continuous-time dynamical system 
\begin{align} \label{eqn dyn_sys}
    \frac{\mathrm{d} \boldsymbol{u}(t)}{\mathrm{d} t} = \boldsymbol{A} \cdot \boldsymbol{u}(t) + \boldsymbol{e}(t) \quad \mbox{with} \quad \boldsymbol{A} \in \mathbb{R}^{m^2},
\end{align}
for the evolution of the solution trajectory $\boldsymbol{u}(t)$ in a reconstructed state space
\begin{align} \label{eqn embeding}
    \boldsymbol{u}(t) \approx \mathcal{E}[u(t)] \quad \mbox{with} \quad \mathcal{E}: \mathbb{R} \to \mathbb{R}^m.
\end{align}
The discrete measurements are parameterized by a continuous and differentiable function $u(t)$. 

The aim of this paper is to propose an algorithmic scheme enabling the reconstruction of the latent continuous dynamics. Through a self-consistent process to be discussed in Sec. \ref{sec: method}, the matrix $\boldsymbol{A}$, the embedding $\mathcal{E}(\cdot)$, and the fitting function $u(t)$ are constructed using deep neural networks, which are learned jointly. The proposed scheme is tested on a synthetic time series, which is sampled from the Lorenz attractor \citep{Lorenz63} and masked by additive Gaussian noise, in Sec. \ref{sec: case_study}. Finally, the limitations of the proposed scheme are discussed in Sec. \ref{sec: limitations}, where conclusions are drawn.

\section{Related works}

\textbf{State Space reconstruction}. State space reconstruction with the time delayed vector was first proposed by Ruelle (cf. \citet{Weigend94}) and \citet{Packard80}, and later proved by \citet{Takens81}. In order to remove noise, \citet{Sauer94} applied a low-pass filter to the delayed vector space. Recently, \citep{Jiang17,Lusch18,Gilpin20,Ouala20} have shown that the state space reconstruction from noisy data can be achieved using an autoencoder as the embedding function $\mathcal{E}(\cdot)$. \citet{Gilpin20} incorporated the false-nearest-neighbor (FNN) algorithm \citep{Kennel92} into a loss function to penalize the encoder outputs in the redundant dimensions. As a consequence, the reconstructed attractor is confined to a subspace smaller than the configuration space. However, by incorporating the FNN algorithm into the loss function, one penalizes the dynamics not only in the redundant dimensions, but also on the solution manifold, leading to a dimensionality collapse of the reconstructed attractor reported in \citep{Gilpin20}.

\textbf{Neural ordinary differential equation (ODE)}. The canonical approach for learning neural ODEs from data uses the adjoint method and calls for numerical solvers \citep{Chen2018,Ayed19,Rackauckas20}. Depending on the selected numerical schemes, the differential equation (\ref{eqn dyn_sys}) allows for various difference representations, signifying an error. Numerical error accumulates, or even diverges, as one iterates over time, and ultimately affects the loss function. Moreover, in order to obtain the gradients of the loss with respect to network parameters, one needs to solve Eq. (\ref{eqn dyn_sys}) forwards and a corresponding adjoint ODE backwards, at each iteration. This can be computationally prohibitive for complex network architectures.

\section{Methods} \label{sec: method}

\subsection{Fitting and filtering} \label{sec: data_imputation}

Time series data are often unevenly spaced. In order to obtain a continuous limit, we parameterize the measured time series using neural networks
\begin{align} \label{eqn u_t}
    u(t) = \mathcal{N}_u(t;\boldsymbol{\theta}_u), \quad \mathcal{N}_u(\cdot): \mathbb{R} \to 
    \mathbb{R}.
\end{align}
Let $s_N = [s(1), s(2), ..., s(N)]$ be a segment of the time series measured at times $[t_1, t_2, ..., t_N]$, respectively. The deviation loss associated with the fitting which reads
\begin{align}
    L_\text{fit} = \frac{1}{N \sigma_s^2} \sum_{i=1}^{N} \left[ s(i) - \mathcal{N}_u(t_i; \boldsymbol{\theta}_u) \right]^2,
\end{align}
is normalized by the batch variance, $\sigma_s^2 = \frac{1}{N} \sum_{i=1}^N \left[ s(i) - \text{mean}(s_N) \right]^2$. 

For noisy datasets, a direct minimization of $L_\text{fit}$ with respect to $\boldsymbol{\theta}_u$ will lead to an overfitting. Assuming that the dynamics of $u(t)$ is characterized by the underlying neural ODE (\ref{eqn dyn_sys}), we include the deviation $\boldsymbol{e}(t)$ as a regularizer
\begin{align}
    L_\text{ode} = \frac{1}{Md} \sum_{i=1}^{M} \sum_{j=1}^{m} \left[ e_j(t_i) \right]^2,
\end{align}
where the embedding dimension, $d \leq m$, is learned during the training. The fitting process is self consistent as the error vector $\boldsymbol{e}(t) \in \mathbb{R}^m$ is an implicit function of $\boldsymbol{\theta}_u$. Moreover, we augment the data by randomly sampling $M$ points from $u(t)$ for $t \in [t_1 + (m-1)\tau, t_N]$ at each iteration, where the interval $(m-1)\tau$ stems from the delay vector. A successive re-sampling covers the entire solution manifold of Eq. (\ref{eqn dyn_sys}). We defer our discussion on the learning algorithm for $\mathcal{E}(\cdot)$ and $d$ to Sec. \ref{sec: state_space_reconstruction} and on the functional form of $\boldsymbol{e}(t)$ to Sec. \ref{sec: neural ODE}, respectively. Therefore, the total loss function for $\mathcal{N}_u(\cdot)$ is
\begin{align} \label{eqn L_u}
    L_{u} = L_\text{fit} + \lambda_u L_\text{ode},
\end{align}
where the weight $\lambda_u$ is a hyper-parameter. A joint minimization of $L_\text{fit}$ and $L_\text{ode}$ ensures the smoothness of the solution trajectory and filters out additive noise.

\subsection{State space reconstruction} \label{sec: state_space_reconstruction}

Adopting the idea from \citet{Gilpin20}, we start with a reasonably large value $m$ for the configuration space, $\mathbb{R}^m$, where Eq. (\ref{eqn dyn_sys}) lives, and search for an embedding subspace $\mathbb{R}^d$, with $d \leq m$, which contains the attractor. Given $u(t)$, the delay vector can be expressed as
\begin{align}
    \tilde{\boldsymbol{u}}(t)^T = [u(t), u(t-\tau), ..., u(t - (m-1)\tau) ].
\end{align}
To treat the redundancy associated with large values of $m$,  we calculate the fraction of false nearest neighbors $\gamma$, a heuristic first proposed by \citet{Kennel92}. False neighbors of a trajectory point in too low an embedding dimension will separate as the embedding dimension increases until all neighbors are real. Therefore, an appropriate embedding dimension $d$ can be inferred by examining how $\gamma$ varies as a function of dimension. See supplementary material for details.

Let $\boldsymbol{\gamma} = [\gamma_1, ..., \gamma_m]$ be a list containing the fraction of false nearest neighbors for each subspace in $\mathbb{R}^m$. Instead of incorporating $\boldsymbol{\gamma}$ into a loss function \citep{Gilpin20}, we introduce a binary mask 
\begin{align} \label{eqn weight_fnn}
    \boldsymbol{w}^T = \lceil \text{relu}(\boldsymbol{\gamma} - \epsilon) \rceil = [\underbrace{1, ..., 1}_{d}, \underbrace{0, ..., 0}_{m-d}],
\end{align}
which decomposes the configuration space $\mathbb{R}^m$ into an embedding $\mathbb{R}^d$ and $(m-d)$ redundant dimensions. Here, $\lceil \cdot \rceil$ denotes the ceiling function. Following \citet{Kennel92}, we take $\epsilon = 0.01$.

Let us consider a state space reconstruction using either the method of delay where the state vector is:
\begin{align} \label{eqn mod}
    \boldsymbol{u} = \boldsymbol{w}^T \tilde{\boldsymbol{u}},
\end{align}
or the autoencoder where the bottleneck is created by the mask $\boldsymbol{w}^T$
\begin{subequations} \label{eqn autoencoder}
\begin{align}
    \mbox{Encoder}: &\quad \boldsymbol{u}(t) = \boldsymbol{w}^T \mathcal{N}_e[\tilde{\boldsymbol{u}}(t); \boldsymbol{\theta}_e], \quad \mathcal{N}_e(\cdot): \mathbb{R}^m \to \mathbb{R}^m, \label{eqn encoder} \\
    \mbox{Decoder}: &\quad \hat{\boldsymbol{u}}(t) = \mathcal{N}_d[\boldsymbol{u}(t); \boldsymbol{\theta}_d], \quad \mathcal{N}_d(\cdot): \mathbb{R}^m \to \mathbb{R}^m. \label{eqn decoder}
\end{align}
\end{subequations}
The inclusion of $\boldsymbol{w}^T$ compresses the outputs of $\mathcal{N}_e(\cdot)$ to the embedding, while the decoder ensures information conservation. The associated reconstruction loss is
\begin{align} \label{eqn L_rec}
    L_\text{rec} = \frac{d}{M m \sum_{i=1}^{d}\sigma_{u_i}^2} \sum_{i=1}^M \sum_{j=1}^{m} \left[ \hat{u}_j(t_i) - \tilde{u}_j(t_i) \right]^2,
\end{align}
where $\sigma_{\boldsymbol{u}}$ denotes the standard deviation of $\boldsymbol{u}$ in the batch direction. To enforce an isotropic expansion of the attractor, we consider minimizing the following loss function 
\begin{align} \label{eqn L_pca}
    L_\text{exp} = \frac{2}{d(d-1)} \sum_{i=1}^{d-1} \sum_{j=i+1}^{d}  K_{u_iu_j}^2 + \frac{1}{d} \sum_{i=1}^d \left[ \sigma_{u_i} - \text{mean}(\sigma_{\boldsymbol{u}}) \right]^2,
\end{align}
where $\boldsymbol{K}_{\boldsymbol{u}\boldsymbol{u}} \in \mathbb{R}^{m \times m}$ denotes the covariance matrix of $\boldsymbol{u}$, such that the outputs of the encoder span an orthogonal basis, while the second term forces the unfolding to be isotropic. Once more, we include $L_\text{ode}$ as a regularizer. Thus, with the weights $\lambda_{e,1}$ and $\lambda_{e,2}$ being hyper-parameters, the loss functions for the encoder and the decoder are, respectively
\begin{align} \label{eqn L_e}
    L_e = L_\text{rec} + \lambda_{e,1} L_\text{ode} + \lambda_{e,2} L_\text{exp} \quad \mbox{and} \quad L_d = L_\text{rec}.
\end{align}

\subsection{Neural dynamical system} \label{sec: neural ODE}

To confine the dynamics of Eq. (\ref{eqn dyn_sys}) to the $d$-dimensional embedding, we introduce an FNN-informed attention $\boldsymbol{w}\cdot \boldsymbol{w}^T$ to the output matrix of a neural network $\mathcal{N}_f(\cdot): \mathbb{R}^m \to \mathbb{R}^{m^2}$, such that
\begin{align}
    \boldsymbol{A} = (\boldsymbol{w}\cdot \boldsymbol{w}^T) \odot \mathcal{N}_f[\boldsymbol{u}(t); \boldsymbol{\theta}_f] = 
    \begin{bmatrix}
    a_{11} & \dots  & a_{1d} & 0      &\dots  & 0        \notag\\
    \vdots & \ddots & \vdots & \vdots &\ddots & \vdots   \notag\\
    a_{d1} & \dots  & a_{dd} & 0      &\dots  & 0        \notag\\
    0      & \dots  & 0      & 0      &\dots  & 0        \notag\\
    \vdots & \ddots & \vdots & \vdots &\ddots & \vdots   \notag\\
    0      & \dots  & 0      & 0      &\dots  & 0        \notag\\
    \end{bmatrix}.
\end{align}
 Given $\boldsymbol{u}(t)$ defined in Sec. \ref{sec: state_space_reconstruction}, the deviation vector is given by
\begin{align}
    \boldsymbol{e}(t) = \frac{d \boldsymbol{u}(t)}{\mathrm{d} t} - \boldsymbol{A}\cdot\boldsymbol{u}(t) = \frac{d \boldsymbol{u}(t)}{\mathrm{d} t} - \left\{(\boldsymbol{w}\cdot \boldsymbol{w}^T) \odot \mathcal{N}_f[\boldsymbol{u}(t); \boldsymbol{\theta}_f] \right\} \cdot \boldsymbol{u}(t),
\end{align}
where the time derivative is calculated using auto-differentiation \citep{Baydin17}.

The dimensionality reduction from the configuration space to the embedding one is a common feature of dissipative systems \citep{Temam12}. Thus, we impose that the divergence of the vector field $\boldsymbol{F}(\boldsymbol{u}) = \boldsymbol{A}\cdot \boldsymbol{u}$ is negative. Being a global property of the dynamical system (\ref{eqn dyn_sys}), at each iteration, we sample the configuration space with $M$ points, denoted by $\boldsymbol{u}_s$, leading to a loss function
\begin{align}
    L_\text{div} = \frac{1}{Md} \sum_{i=1}^{M} \left\{ \text{relu}\left[ \text{div} \boldsymbol{F}(\boldsymbol{u}_s(t_i)) \right] \right\}^2.
\end{align}

With $\lambda_f$ being a hyper-parameter, a joint minimization of $L_\text{ode}$ and $L_\text{div}$ 
\begin{align}
    L_{f} = L_\text{ode} + \lambda_{f} L_\text{div},
\end{align}
enforces the long-term dynamics of solution trajectories on the attractor.

\subsection{Algorithm}

The schematic algorithm \ref{algo_1} shown below brings together all pieces introduced in the previous sections. For neural dynamical systems whose state vector is reconstructed using the method of delay (\ref{eqn mod}), one mitigates the need for learning an autoencoder (\ref{eqn autoencoder}).

\begin{algorithm}
  \caption{Joint learning for neural dynamical systems with autoencoder}\label{algo_1}
  \textbf{Input:} Mini-batched training samples $\{ s_{N}[1], s_{N}[2], ..., s_{N}[S] \}$ and the corresponding time labels. \\
  Guess initial parameters $\{ \boldsymbol{\theta}_u[1], \boldsymbol{\theta}_u[2],..., \boldsymbol{\theta}_u[S] \}$, $\boldsymbol{\theta}_e$, $\boldsymbol{\theta}_d$ $\boldsymbol{\theta}_f$ and initialize $\boldsymbol{\gamma} = \boldsymbol{w}^T = [1,...,1]$.
  \begin{algorithmic}[1]
  \While{not converged}{ 
  \State Create an empty list: $U$.
  \For{$i = 1,...,S$}
  \State Randomly draw $M$ samples from $u(t)[i]$ and append to list $U$.  
  \State Randomly draw $M$ samples from the configuration space$^\dagger$.
  \State Compute loss functions: $L_\text{fit}$, $L_\text{rec}$, $L_\text{ode}$ and $L_\text{div}$.
  \State Optimize $\boldsymbol{\theta}_u[i]$ using gradients: $\partial_{\boldsymbol{\theta}_u[i]} L_\text{fit} + \lambda_u \partial_{\boldsymbol{\theta}_u[i]} L_\text{ode}$. 
  \State Optimize $\boldsymbol{\theta}_e$ using gradients: $\partial_{\boldsymbol{\theta}_e} L_\text{rec} + \lambda_{e,1} \partial_{\boldsymbol{\theta}_e} L_\text{ode} + \lambda_{e,2} \partial_{\boldsymbol{\theta}_e} L_\text{exp}$.
  \State Optimize $\boldsymbol{\theta}_d$ using gradients: $\partial_{\boldsymbol{\theta}_d} L_\text{rec}$.
  \State Optimize $\boldsymbol{\theta}_f$ using gradients: $\partial_{\boldsymbol{\theta}_f} L_\text{ode} + \lambda_f \partial_{\boldsymbol{\theta}_f} L_\text{div}$.
  \EndFor
  \State Compute epoch-wise $\hat{\boldsymbol{\gamma}}$ using $U$ and update $\boldsymbol{\gamma}$ using a moving average: $\boldsymbol{\gamma} = (1-\alpha) \boldsymbol{\gamma} + \alpha \hat{\boldsymbol{\gamma}}$.
  \State Compute and update $\boldsymbol{w}^T$ using Eq. (\ref{eqn weight_fnn}).
  \EndWhile}
  \end{algorithmic}
  \hspace*{\algorithmicindent} \textbf{Output}: Optimized parameters $\{ \boldsymbol{\theta}_u[1], \boldsymbol{\theta}_u[2],..., \boldsymbol{\theta}_u[S] \}$, $\boldsymbol{\theta}_e$, $\boldsymbol{\theta}_e$, $\boldsymbol{\theta}_f$ and the weights $\boldsymbol{w}^T$. \\
  Hyper-parameters in this paper: $S=128$, $M=64$, $\lambda_u = \lambda_{e,1} = \lambda_{e,2} = \lambda_f = 1$ and $\alpha = 0.1$.\\
  $^\dagger$ The configuration space is confined by the output activation function of the encoder, e.g. $\tanh$.
\end{algorithm}

\section{Experiments with synthetic time series} \label{sec: case_study}

\begin{figure}
   \centering
   \noindent\makebox[\textwidth]{
   \includegraphics[width=\textwidth]{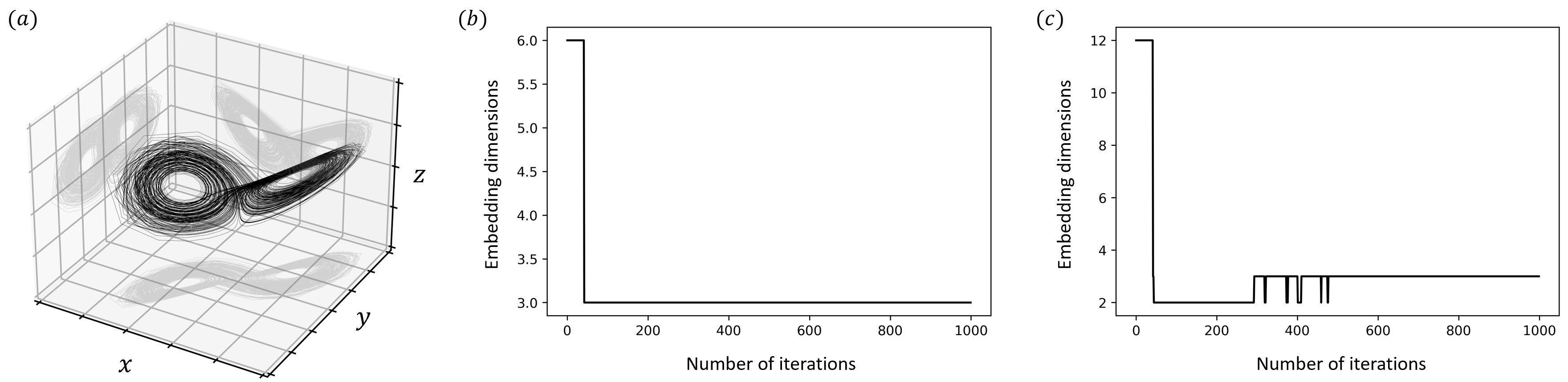}}
   \caption{\label{fig:delay_dim} $(a)$ Lorenz attractor. Dimension $d=3$ obtained from the FNN method for embedding in an $(b)$ $m=6$; and an $(c)$ $m=12$ dimensional configuration space over $1000$ iterations.}
\end{figure}

State space reconstruction \citep{Sauer94,Jiang17,Ouala20,Gilpin20} and recovered neural ODEs \citep{Chen2018} are commonly assessed by visualizing the reconstructed attractor and by forecasting the continuation of the time series. In all herein cited works, synthetic time series generated by solving the \citet{Lorenz63} equations 
\begin{align}\label{eqn Lorenz}
    \frac{d x}{d t} = \sigma(y-x), \quad
    \frac{\mathrm{d} y}{\mathrm{d} t} = x(\rho - z) - y, \quad
    \frac{\mathrm{d} z}{\mathrm{d} t} = xy - \beta z,
\end{align}
with $\sigma = 10$, $\rho = 28$, $\beta = 8/3$, were considered. Our dataset is the $x$-coordinate of a trajectory generated by solving Eqs. (\ref{eqn Lorenz}) for $10,200$ time steps from one initial condition $(x,y,z) = (0, 1, 1.05)$. With a step size $\Delta t = 0.05$, a visualization of the Lorenz attractor is shown in Fig. \ref{fig:delay_dim} $(a)$, where the time series $t\in[0,510]$ completes one oscillation every $15$ to $20$ time steps. In addition, the time series are masked by an Gaussian white noise, with $\eta$ being a ratio:
\begin{align} \label{eqn noise}
    s'[k] = s[k] + N(0, \sigma^2) \quad \mbox{where} \quad \sigma = \eta \sigma_s.
\end{align}

In the model, we employ a residual connection \citep{He16} around two sub-layers made of $32$ neurons regularized with batch normalization \citep{Ioffe15} and activated by a $\tanh$ function. Networks $\mathcal{N}_u, \mathcal{N}_e, \mathcal{N}_d$ and $\mathcal{N}_f$ consist of a stacking of $3$ and $5$ residual blocks, respectively. By rescaling the training dataset to a range $[-1,1]$, the output activation functions for $\mathcal{N}_u, \mathcal{N}_e, \mathcal{N}_d$ are $\tanh$. Since there is no a priori information on $\boldsymbol{A}$ (except for the divergence of $\boldsymbol{A}\cdot\boldsymbol{u}$), we use a linear activation for the output of $\mathcal{N}_f$. Then, we corrupt the input of the encoder with Gaussian noise $N(0, 0.5^2)$ as in \citep{Vincent2008a,Vincent10,Gilpin20}; to reduce overfitting, we apply a dropout regularization \citep{Srivastava14} with a rate $0.1$ just before each network's output layer.

We separated the time series into a training $t \in [0,490]$, a validation $t \in [490, 500]$ and a test $t\in[500, 510]$ set; the training set was divided into $128$ batches with an $(m-1)\tau$ overlapping between each batch. The FNN-algorithm converged to and remained on a $d=3$ embedding during the training, cf. Fig. \ref{fig:delay_dim} $(b)$ and $(c)$. Without loss of generality, we selected $m = 6$ and $\tau = 0.1$. We trained our model using the ADAM optimizer \cite{Kingma15} and on a NVIDIA 2080Ti GPU; iterating over an epoch took around $2$ and $4$ seconds without and with the autoencoder, respectively. Thus, we pretrained the models $\mathcal{N}_u$ and $\mathcal{N}_f$ with $\boldsymbol{u}$ reconstructed using the method of delay (\ref{eqn mod}) for $15,000$ iterations, and then turn on the autoencoder (\ref{eqn autoencoder}) for a fine tuning for $15,000$ iterations. The results were compared with a direct training using the method of delay for $30,000$ iterations. We retained the models with the lowest normalized mean square error (NMSE) on the validation set
\begin{align} \label{eqn NMSE}
    \text{NMSE} = \sum_{i=1}^{N}(\text{truth}_i - \text{prediction}_i)^2 / \sum_{i=1}^{N} \left[\text{truth}_i - \text{mean}(\text{truth})\right]^2.
\end{align}

\subsection{Visualization of the latent attractor} \label{sec. vis_attractor}

\begin{figure}
   \centering
   \noindent\makebox[0.85\textwidth]{
   \includegraphics[width=0.85\textwidth]{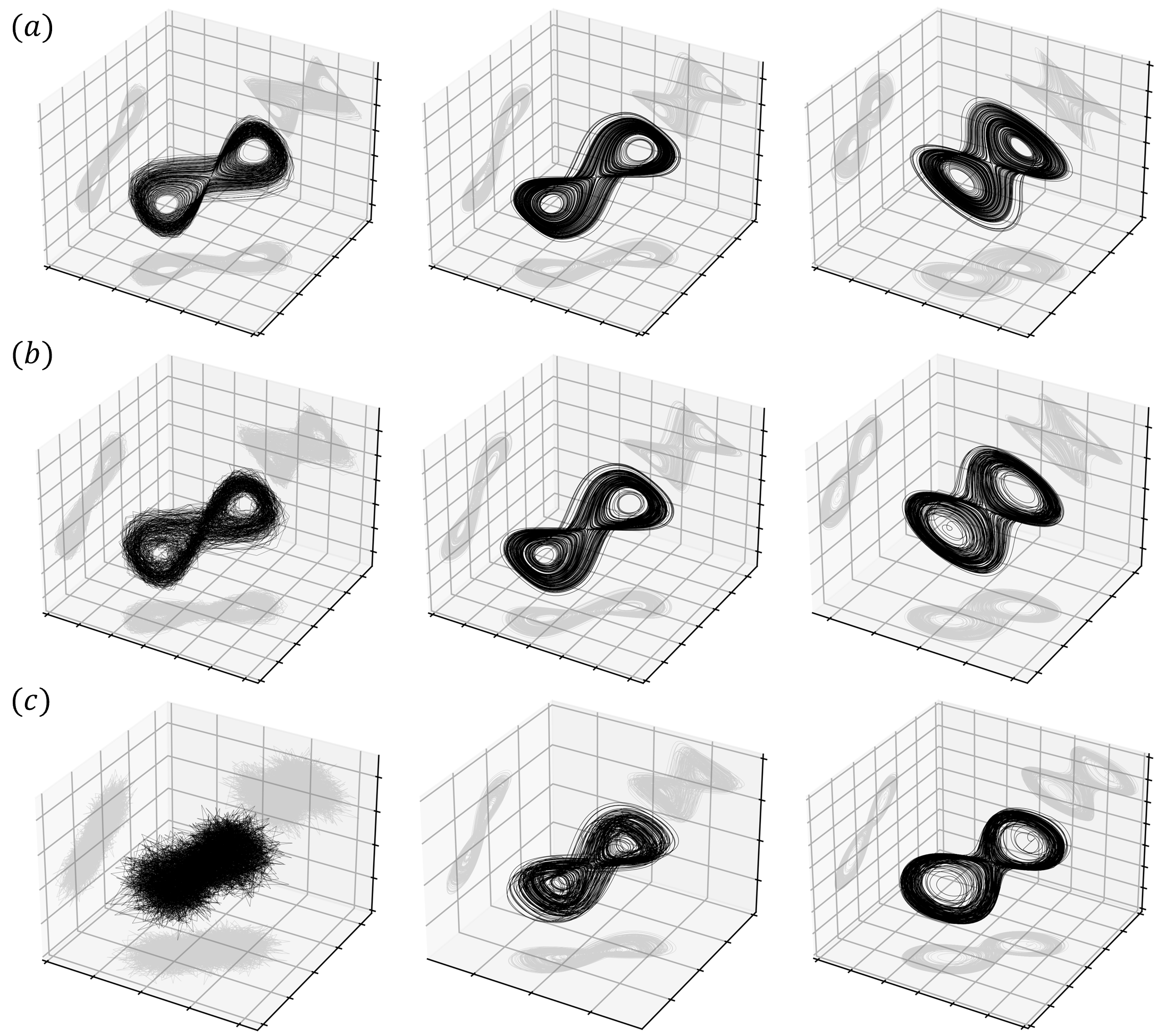}}
   \caption{\label{fig:latent_attractor} Delay attractors reconstructed from noisy measurements (Left) or the parameterization $u(t)$ (Middle). Right: latent attractors reconstructed using the autoencoder. The time series were masked by Gaussian noise with noise ratio $(a)$ $\eta=0$; $(b)$ $\eta=0.15$; and $(c)$ $\eta=0.3$, cf. Eq. (\ref{eqn noise}) }
\end{figure}

Fig. \ref{fig:latent_attractor} shows the reconstructed attractors from noisy measurements. Even for the case $\eta = 0.3$, the $L_\text{ODE}$ regularizer enables us to recover an attractor resembling that observed in noise-free conditions. Note that the filtering process can occur at two stages. First, the regularized parameterization removes, in principle, outliers that are not governed by the latent ODEs (\ref{eqn dyn_sys}). This features a physics-informed filtering, despite the underlying physics is not known a priori, but it is learned in parallel with the filtering process. Second, since the reconstruction loss, cf. Eq. (\ref{eqn L_rec}), is not exactly zero at the end of the training, the deviation of the decoded signal from $u(t)$ may contribute to another filtering process. In order to assess the filtering ability, we compare in Table \ref{table:filtering} the NMSE of the noisy signal, the learned $u(t)$, and the output of the decoder with the noise-free Lorenz time series. An overall $80\%$ noise reduction has been achieved; however, such a reduction is mainly due the inclusion of $L_\text{ode}$ during the trajectory fitting. 

\begin{table}[ht]
\caption{Normalized mean square error of the two-stage filtering process}
\centering 
\begin{tabular}{c c c c}
\hline\hline \\ [0.01ex]  
Case & Raw measurements & $u(t)$ & $\mathcal{N}_d\{\mathcal{N}_e[\tilde{\boldsymbol{u}}(t)]\}$ \\ [1.5ex]
\hline \\  [0.01ex] 
$\eta = 0.00$ & $0.0$                 &  $5.8 \times 10^{-5}$  &  $9.6 \times 10^{-5}$   \\
$\eta = 0.15$ & $2.3 \times 10^{-2}$  &  $2.6 \times 10^{-3}$  &  $2.6 \times 10^{-3}$   \\
$\eta = 0.30$ & $9.1 \times 10^{-2}$  &  $1.6 \times 10^{-2}$  &  $1.3 \times 10^{-2} $  \\ [1ex]   
\hline
\end{tabular}
\label{table:filtering}
\end{table}

For the Lorenz system, each time the signal $x$ goes to zero, the trajectory passes through a sensitive region, i.e. the center, of the attractor where the signal is susceptible to error \citep{Sauer94,Dubois20,Brunton17}. Compared with delay attractors and works using autoencoders \citep{Jiang17,Gilpin20}, the $L_\text{ODE}$-regularized encoder unfolds the central region of the attractor, which may be the reason for the enhanced forecasting horizon discussed in Sec. \ref{sec: direct_continuation}.

\subsection{Continuation of the training time series} \label{sec: direct_continuation}

\begin{figure}
   \centering
   \noindent\makebox[\textwidth]{
   \includegraphics[width=\textwidth]{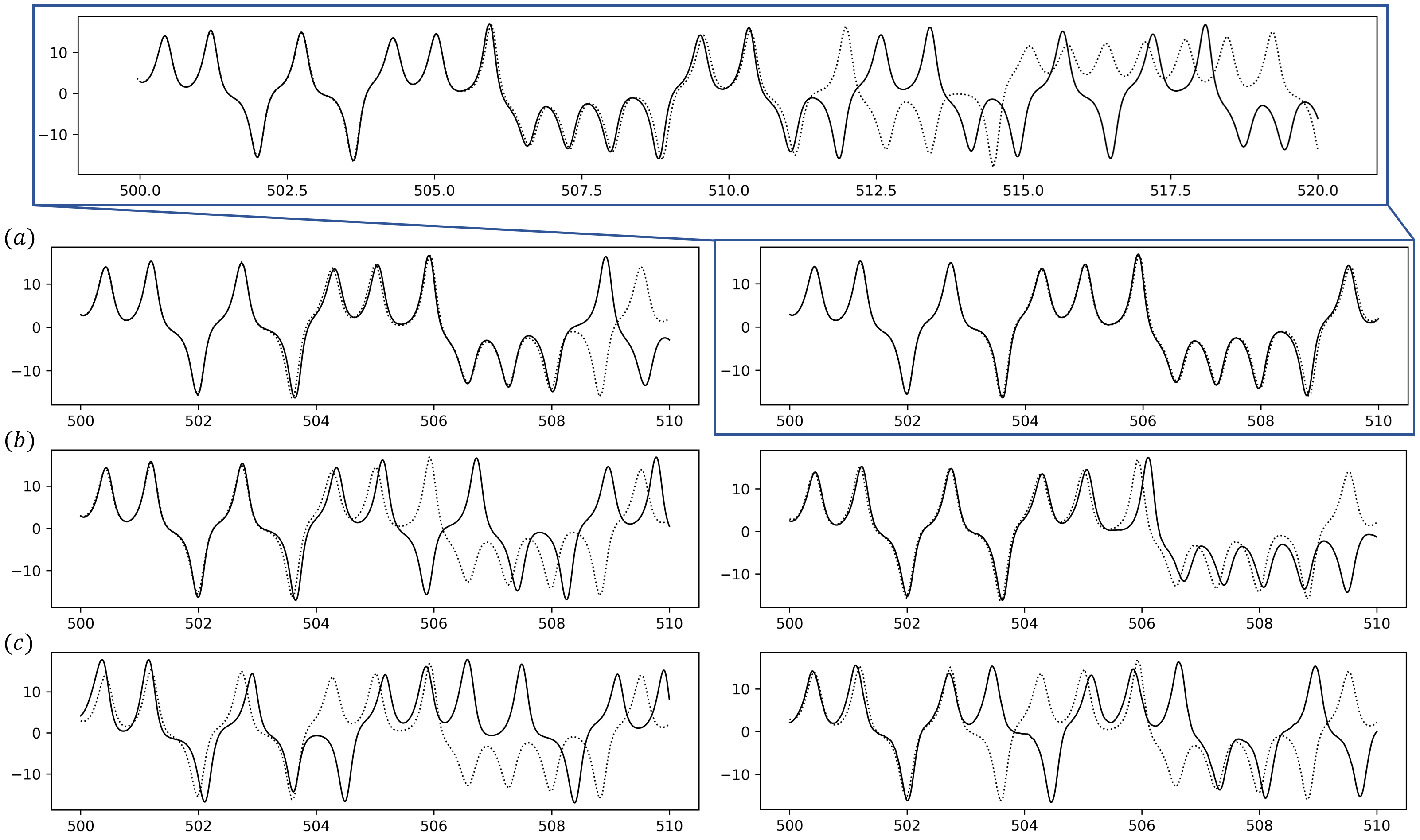}}
   \caption{\label{fig:direct_continuation} Time series prediction for signal from Lorenz attractor masked by an additive Gaussian noise with $(a)$ $\eta = 0$; $(b)$ $\eta = 0.15$; and $(c)$ $\eta = 0.3$ without (left) and with (right) the autoencoder. Solid curve denotes the predicted time series and dashed curve is the true time series. The inset reveals that with the autoencoder and a clean training dataset $\eta=0$, the prediction stays near the correct signal for around $220$ time steps with $\Delta t = 0.05$.}
\end{figure}

Two neural dynamical systems were identified in the state spaces reconstructed either with method of delay (\ref{eqn mod}) or with the autoencoder (\ref{eqn autoencoder}). For both systems, we first solved Eq. (\ref{eqn dyn_sys}) with given initial conditions $\boldsymbol{u}(t_0)$ in the configuration space, and then convert the solutions to the measurement space. With the method of the delay, the conversion reduces to an identity operation; whereas with the autoencoder, the conversion is made using the decoder (\ref{eqn decoder}). Taken $t_0 = 500$, we predict the continuation of the time series for the next $200$ time steps with $\Delta t = 0.05$, and compare both predictions with the exact solution of the Lorenz model, see Fig. \ref{fig:direct_continuation}. Our best result was obtained when integrating the neural dynamical system reconstructed in a latent state space and trained on a noise-free dataset. As Fig. \ref{fig:direct_continuation} shows, the predicted continuation remains close to the exact signal up to $11$ Lorenz time, which is at least $2$ times longer than previously achieved in \citep{Sauer94,Chen2018,Dubois20,Gilpin20} and on par with \citet{Ouala20} who implemented a neural autoregression model, cf. Eq. (\ref{eqn NAR}) in the latent space. The inclusion of noise leads to an deterioration in the prediction horizon, as expected. With increasing noise ratio, our algorithm fails finding the precise initial conditions. The sensitive dependence on initial conditions of chaotic dynamical systems prohibits forecasting. 

\begin{figure}
   \centering
   \noindent\makebox[\textwidth]{
   \includegraphics[width=\textwidth]{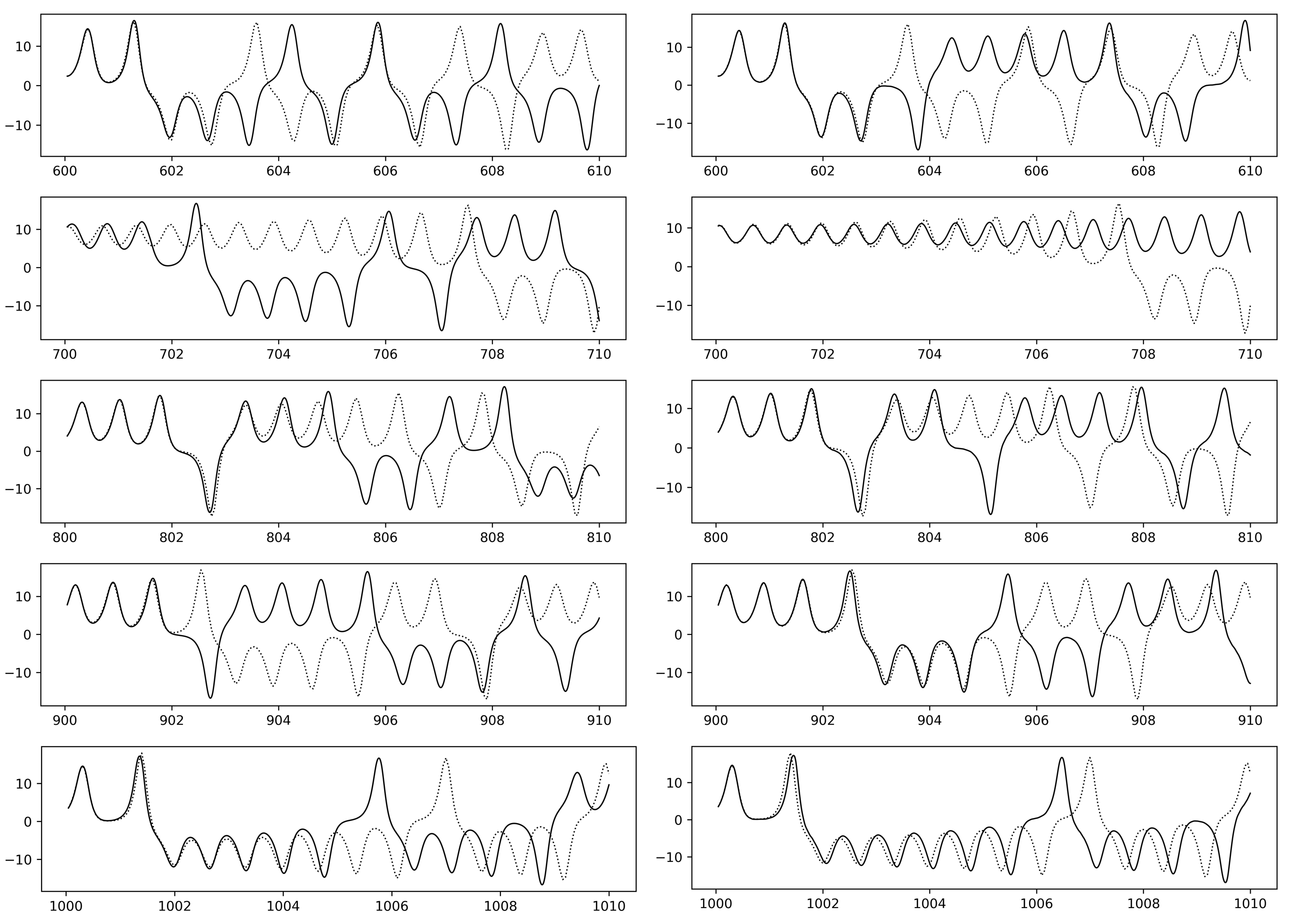}}
   \caption{\label{fig:future_continuation} Prediction for five successive continuations of the training dataset $t\in[0, 490]$ further into the future without (left) and with (right) the autoencoder. The solid curve is the predicted continuation and the dashed curve denotes the ground truth.}
\end{figure}

Fig. \ref{fig:future_continuation} shows continuations of the noise-free training dataset $t\in[0, 490]$ into the future. The initial condition at, e.g. $t_0=600$, is determined from measurements in an interval $t \in [599,601]$ by minimizing $L_{u}$, cf. Eq. (\ref{eqn L_u}). The parameters $\boldsymbol{\theta}_f$, $\boldsymbol{\theta}_e$ and $\boldsymbol{\theta}_d$ are fixed during the inference phase. Despite a degradation of prediction horizon for all cases, a better performance is achieved for state space reconstruction using the autoencoder, coming with the cost of doubled computation time.

\subsection{Forecasting from different initial conditions}

\begin{figure}
   \centering
   \noindent\makebox[\textwidth]{
   \includegraphics[width=\textwidth]{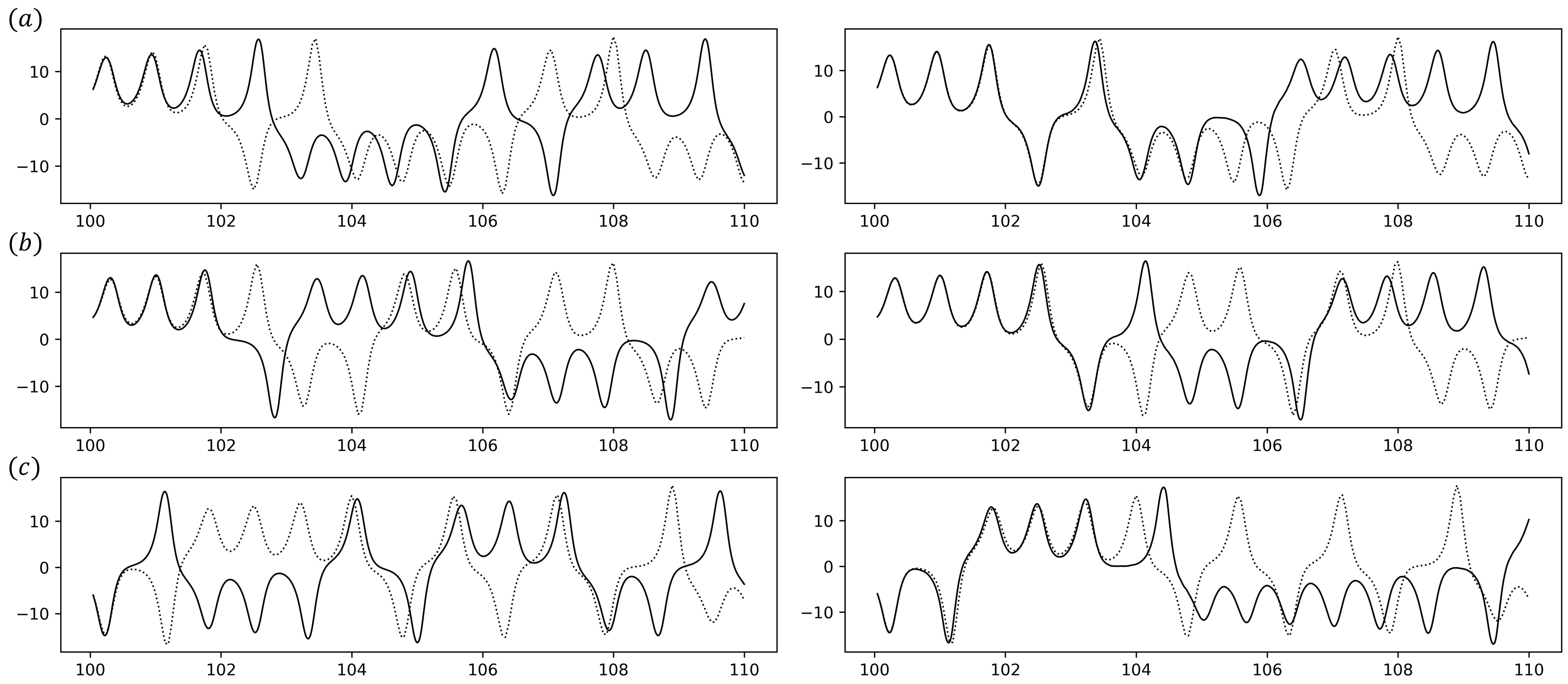}}
   \caption{\label{fig:prediction_different} Prediction $(t\in[100,110])$ for Lorenz time series seeded with different initial conditions ($t=0$) from the training dataset $(a)$ $(x,y,z) = (1.02, 0.05, -1.67)$; $(b)$ $(x,y,z) = (3.14, -1.59, 2.65)$; and  $(c)$ $(x,y,z) = (2.00, 3.00, 4.25)$ without (left) and with (right) the autoencoder. Solid curve denotes the predicted time series and dashed curve is the true time series.}
\end{figure}

To assess the transferability of our prediction model, we study forecasting for time series initiated with different initial conditions in the training dataset. The Lorenz model (\ref{eqn Lorenz}) is integrated with given initial conditions for the period $t \in [0, 110]$, wherein the interval $t \in [99, 101]$ is used to determine the state vector $\boldsymbol{u}(t_0)$ at $t_0=100$. Then, the reconstructed dynamical system is solved for $t\in[100, 110]$. The results are shown in Fig. \ref{fig:prediction_different}. Compared with direct continuations (Figs. \ref{fig:direct_continuation}, \ref{fig:future_continuation}), the deterioration in prediction horizon suggests that the reconstructed dynamical system overfits to the training dataset, which is seeded with a single initial condition. Therefore, should a reconstruction of the universal dynamical system be the goal, one needs to augment the training dataset with time series seeded with different initial conditions. Nevertheless, compared with the method of delay, our results show that state space reconstruction using the regularized autoencoder (\ref{eqn autoencoder}) mitigates the overfitting.

\section{Discussion and conclusions} \label{sec: limitations}

We have introduced a general method which allows us to filter a noisy time series without a priori information on the measurement, to reconstruct an attractor from the filtered results, and to learn a latent dynamical system which underlies the time series in a self-consistent manner. First, the measured time series is parameterized as a continuous function $u(t)$ using deep neural networks. The measurement is then augmented and used to reconstruct the state space using either the method of delay and an autoencoder. The embedding dimension, which creates a bottleneck, is automatically searched using the FNN algorithm during the training. Finally, an approximated dynamical system underlying the temporal evolution serves as a regularizer for $u(t)$ to filter out the noise and for the encoder to determine an optimal embedding, completing the self-consistent cycle.

The proposed framework has been tested by forecasting the continuation of an univariate time series sampled from the Lorenz attractor. To the best of our knowledge, our prediction horizon is significantly longer than most published ones. Moreover, with an additive Gaussian noise, an overall $80\%$ noise reduction is achieved for a chaotic signal. While we adopted the same architecture for all neural networks and keep the regularizer strength constant $\lambda_u = \lambda_{e,1} = \lambda_{e,2} = \lambda_f = 1$, \citet{Gilpin20,Wang21a,Wang21b} recently showed that a careful tuning of the regularizer strength and a selection of the network architecture can speed up the convergence and improve the quality of the learned model to a large degree. Thus, we trust that our results can still be improved. It also remains to test the algorithm on a broader class of synthetic deterministic time series. Eventually, we hope to apply it as an auxiliary tool to explore hidden physics from experimental data and to extract relevant information from direct numerical simulations.

The fundamental assumption underlying this work is that the time series is governed by some unknown ordinary differential equations. Therefore, our algorithm is only applicable to deterministic time series masked by additive noise. Being aware that most time series of practical interest, e.g. financial data, are stochastic, one needs to replace Eq. (\ref{eqn dyn_sys}) by stochastic differential equations and reformulate the loss functions, accordingly.

\section*{Acknowledgment}
The authors would like to thank Cherif Assaf for providing computational resources and Subodh Mhaisalkar for support.

\section*{Appendix} \appendix
\section{Calculation for the fraction of false nearest neighbors} \label{app. fnn}


Consider a random sampling of $M$ points from an $m$-dimensional delay vector to form a batch
\begin{align} \label{eqn delay_matrix}
    \begin{bmatrix}
    u(t_1) & u(t_1 - \tau) & \dots  & u(t_1 - (m-1)\tau) \\
    u(t_2) & u(t_2 - \tau) & \dots  & u(t_2 - (m-1)\tau) \\
    \vdots & \vdots        & \ddots & \vdots             \\
    u(t_M) & u(t_M - \tau) & \dots  & u(t_M - (m-1)\tau)
    \end{bmatrix},
\end{align} 
Let us reorganize Eq. (\ref{eqn delay_matrix}) into the following form: 
\begin{align} \label{eqn matrix}
    \boldsymbol{H} =
    \left[ 
    \begin{bmatrix}
    u(t_1) & 0      &  \dots  & 0 \\
    u(t_2) & 0      &  \dots  & 0 \\
    \vdots & \vdots &  \ddots & 0 \\
    u(t_M) & 0      &  \dots  & 0 
    \end{bmatrix}, 
    \mbox{...},
    \begin{bmatrix}
    u(t_1) & u(t_1 - \tau) & \dots  & u(t_1 - (m-1)\tau) \\
    u(t_2) & u(t_2 - \tau) & \dots  & u(t_2 - (m-1)\tau) \\
    \vdots & \vdots        & \ddots & \vdots             \\
    u(t_M) & u(t_M - \tau) & \dots  & u(t_M - (m-1)\tau)
    \end{bmatrix}
    \right].
\end{align}
wherein the $d$-th component of $\boldsymbol{H}$, denoted by $\boldsymbol{H}_d \in \mathbb{R}^{M \times m}$, represents the batched $d$-dimensional delay vector for $d = 1, ..., m$. The Euclidean length of each delay vectors in the batch $\boldsymbol{H}_d$ is
\begin{align}
    \boldsymbol{L}_d = \left[ \sum_{j=1}^{d} \left( H_d[i,j] \right)^2 \right]^{1/2} \in \mathbb{R}^{M \times 1}.
\end{align}
Using $\boldsymbol{H}[d]$ and $\boldsymbol{L}[d]$, the Euclidean distances between each pairs of two vectors in $\boldsymbol{H}[d]$ can be expressed by the following symmetric matrix
\begin{align}
    \boldsymbol{D}_d = \left[ \boldsymbol{L}_d^2 + \left( \boldsymbol{L}_d^2 \right)^T - 2 \boldsymbol{H}_d \cdot \boldsymbol{H}_d^T \right]^{1/2} \in \mathbb{R}^{M \times M}.
\end{align}
Collecting the nearest neighbors, i.e. the smallest off-diagonal elements, along the horizontal direction
\begin{align}
    R_d[i] = \text{min}\left( D_d[i,j] \right) \quad \mbox{for} \quad j > i, 
\end{align}
the FNN-algorithm considers the element $i$ as a false nearest neighbor if either of the following two tests fails. They are
\begin{subequations}
\begin{align}
    \left[\frac{R_d[i]^2 - R_{d-1}[i]^2}{R_d[i]^2}\right]^{1/2} > R_\text{tol}, \label{eqn is_false_change}\\
    \frac{R_d[i]}{R_A} > A_\text{tol}, \label{eqn is_false_jump}
\end{align}
\end{subequations}
for $d \geq 2$. Here the recommended values for the thresholds in \citet{Kennel92} are $R_\text{tol} = 10$ and $A_\text{tol} = 2$, respectively, and
\begin{align}
    R_A^2 = \frac{1}{M} \sum_{i=1}^{M} \left[u(t_i) - \text{mean}[u(t)] \right]^2.
\end{align}
Denoting 
\begin{align}
    \gamma_d[i] = 
    \begin{cases}
      1, & \text{if Eq. (\ref{eqn is_false_change}) is True or Eq. (\ref{eqn is_false_jump}) is True or $d=1$}, \\
      0, & \text{otherwise},
    \end{cases}
\end{align}
the fraction of the false nearest neighbors associated with $d$-dimensional embedding is
\begin{align}
    \gamma_d = \frac{1}{M} \sum_{i=1}^{M} \gamma_d[i].
\end{align}
Therefore, the fraction of false nearest neighbors associated with each sub-dimension $d \leq m$ is 
\begin{align}
    \gamma = [\gamma_1, \gamma_2, ..., \gamma_m].
\end{align}

\bibliographystyle{unsrtnat}

\begin{thebibliography}{1}

\bibitem[Yule(1927)]{Yule27}
G.~U. Yule.
\newblock Vii. on a method of investigating periodicities disturbed series,
  with special reference to wolfer's sunspot numbers.
\newblock \emph{Philosophical Transactions of the Royal Society of London.
  Series A}, 226:\penalty0 267--298, 1927.

\bibitem[Lapedes and Farber(1987)]{Lapedes87}
A.~Lapedes and R.~Farber.
\newblock Nonlinear signal processing using neural networks: Prediction and
  system modelling.
\newblock Technical report, Los Alamos National Laboratory, United States,
  1987.

\bibitem[Weigend et~al.(1990)Weigend, Huberman, and Rumelhart]{Weigend90}
A.~S. Weigend, B.~A. Huberman, and D.~E. Rumelhart.
\newblock Predicting the future: A connectionist approach.
\newblock \emph{International journal of neural systems}, 1\penalty0
  (03):\penalty0 193--209, 1990.

\bibitem[Weigend(1991)]{Weigend91}
A.~S. Weigend.
\newblock \emph{Connectionist architectures for time series prediction of
  dynamical systems}.
\newblock PhD thesis, Stanford University, 1991.

\bibitem[Wan(1994)]{Wan94}
E.~A. Wan.
\newblock Time series prediction by using a connectionist network with internal
  delay lines.
\newblock In \emph{Time Series Prediction: forecasting the future and
  understanding the past}, pages 195--217, 1994.

\bibitem[Saad et~al.(1998)Saad, Prokhorov, and Wunsch]{Wunsch98}
E.~W. Saad, D.~V. Prokhorov, and D.~C. Wunsch.
\newblock Comparative study of stock trend prediction using time
  delay,recurrent and probabilistic neural networks.
\newblock \emph{IEEE Transactions on Neural Networks}, 9\penalty0 (6):\penalty0
  1456--1470, 1998.

\bibitem[Borovykh et~al.(2017)Borovykh, Bohte, and Oosterlee]{Borovykh17}
A.~Borovykh, S.~Bohte, and C.~W. Oosterlee.
\newblock Conditional time series forecasting with convolutional neural
  networks.
\newblock arXiv preprint arXiv:1703.04691, 2017.

\bibitem[Gers et~al.(2002)Gers, Eck, and Schmidhuber]{Gers02}
F.~A. Gers, D.~Eck, and J.~Schmidhuber.
\newblock Applying {LSTM} to time series predictable through time-window
  approaches.
\newblock In \emph{Neural Nets WIRN Vietri-01}, pages 193--200. Springer,
  London, 2002.

\bibitem[Mirowski and LeCun(2009)]{LeCun09}
P.~Mirowski and Y.~LeCun.
\newblock Dynamic factor graphs for time series modeling.
\newblock In \emph{Joint European Conference on Machine Learning and Knowledge
  Discovery in Databases}. Springer, Berlin, Heidelberg, 2009.

\bibitem[Dubois et~al.(2020)Dubois, Gomez, Planckaert, and Perret]{Dubois20}
P.~Dubois, T.~Gomez, L.~Planckaert, and L.~Perret.
\newblock Data-driven predictions of the lorenz system.
\newblock \emph{Physica D: Nonlinear Phenomena}, 408:\penalty0 132495, 2020.

\bibitem[{S. Li et al.}(2019)]{Li19}
{S. Li et al.}
\newblock Enhancing the locality and breaking the memory ottleneck of
  transformer on time series forecasting.
\newblock In \emph{Advances in NeuralInformation Processing Systems (NeurIPS)},
  2019.

\bibitem[Lim et~al.(2021)Lim, Loeff, Arik, and Pfister]{Lim2021}
B.~Lim, N.~Loeff, S.~Arik, and T.~Pfister.
\newblock Temporal fusion transformers for interpretable multi-horizon time
  series forecasting.
\newblock \emph{International Journal of Forecasting}, 2021.

\bibitem[Gorban and Wunsch(1998)]{Gorban98}
A.~N. Gorban and D.~C. Wunsch.
\newblock The general approximation theorem.
\newblock In \emph{Proceedings of the International Joint Conference on Neural
  Networks}, 1998.

\bibitem[Winkler and Le(2017)]{Winkler17}
D.~A. Winkler and T.~C. Le.
\newblock Performance of deep and shallow neural networks, the universal
  approximation theorem, activity cliffs, and qsar.
\newblock \emph{Molecular Informatics}, 36\penalty0 (1-2):\penalty0 1600118,
  2017.

\bibitem[Lin and Jegelka(2018)]{Lin18}
H.~Lin and S.~Jegelka.
\newblock Resnet with one-neuron hidden layers is a universal approximator.
\newblock In \emph{Advances in Neural Information Processing Systems}, 2018.

\bibitem[Takens(1981)]{Takens81}
F.~Takens.
\newblock Detecting strange attractors in turbulence." dynamical systems and
  turbulence.
\newblock In \emph{Dynamical systems and turbulence}, pages 366--381. Springer,
  Berlin, Heidelberg, 1981.

\bibitem[Chen et~al.(2018)Chen, Rubanova, Bettencourt, and Duvenaud]{Chen2018}
R.~T. Chen, Y.~Rubanova, J.~Bettencourt, and D.~Duvenaud.
\newblock Neural ordinary differential equations.
\newblock arXiv preprint arXiv:1806.07366, 2018.

\bibitem[Ayed et~al.(2019)Ayed, de~B{\'e}zenac, Pajot, Brajard, and
  Gallinari]{Ayed19}
I.~Ayed, E.~de~B{\'e}zenac, A.~Pajot, J.~Brajard, and P.~Gallinari.
\newblock Learning dynamical systems from partial observations.
\newblock arXiv preprint arXiv:1902.11136, 2019.

\bibitem[Raissi et~al.(2019)Raissi, Perdikaris, and Karniadakis]{Rassi19}
M.~Raissi, P.~Perdikaris, and G.~E. Karniadakis.
\newblock Physics-informed neural networks: A deep learning framework for
  solving forward and inverse problems involving nonlinear partial differential
  equations.
\newblock \emph{Journal of Computational Physics}, 378:\penalty0 686--707,
  2019.

\bibitem[Rackauckas et~al.(2020)Rackauckas, Ma, Martensen, Warner, Zubov,
  Supekar, Skinner, Ramadhan, and Edelman]{Rackauckas20}
C.~Rackauckas, Y.~Ma, J.~Martensen, C.~Warner, K.~Zubov, R.~Supekar,
  D.~Skinner, A.~Ramadhan, and A.~Edelman.
\newblock Universal differential equations for scientific machine learning.
\newblock \emph{arXiv preprint arXiv:2001.04385}, 2020.

\bibitem[Lorenz(1963)]{Lorenz63}
E.~N. Lorenz.
\newblock Deterministic nonperiodic flow.
\newblock \emph{Journal of atmospheric sciences}, 20\penalty0 (2):\penalty0
  130--141, 1963.

\bibitem[Weigend and Gershenfeld(1994)]{Weigend94}
A.~S. Weigend and N.~A. Gershenfeld, editors.
\newblock \emph{Time Series Prediction: forecasting the future and
  understanding the past}.
\newblock Addison-Wesley, 1994.

\bibitem[Packard et~al.(1980)Packard, Crutchfield, Farmer, and Shaw]{Packard80}
N.~H. Packard, J.~P. Crutchfield, J.~D. Farmer, and R.~S. Shaw.
\newblock Geometry from a time series.
\newblock \emph{Physical review letters}, 45\penalty0 (9):\penalty0 712--716,
  1980.

\bibitem[Sauer(1994)]{Sauer94}
T.~Sauer.
\newblock Time series prediction by using delay coordinate embedding.
\newblock In \emph{Time Series Prediction: forecasting the future and
  understanding the past}, 1994.

\bibitem[Jiang and He(2017)]{Jiang17}
H.~Jiang and H.~He.
\newblock State space reconstruction from noisy nonlinear time series: An
  autoencoderbasedapproach.
\newblock In \emph{International Joint Conference on Neural Networks (IJCNN)},
  2017.

\bibitem[Lusch et~al.(2018)Lusch, Kutz, and Brunton]{Lusch18}
B.~Lusch, J.~N. Kutz, and S.~L. Brunton.
\newblock Deep learning for universal linear embeddings ofnonlinear dynamics.
\newblock \emph{Nature communications}, 9:\penalty0 1--10, 2018.

\bibitem[Gilpin(2020)]{Gilpin20}
W.~Gilpin.
\newblock Deep reconstruction of strange attractors from time series.
\newblock In \emph{Advances in Neural Information Processing Systems
  (NeurIPS)}, 2020.

\bibitem[Ouala et~al.(2020)Ouala, Nguyen, Drumetz, Chapron, Pascual, Collard,
  Gaultier, and Fablet]{Ouala20}
S.~Ouala, D.~Nguyen, L.~Drumetz, B.~Chapron, A.~Pascual, F.~Collard,
  L.~Gaultier, and R.~Fablet.
\newblock Learning latent dynamics for partially-observed chaotic systems.
\newblock \emph{Chaos: An Interdisciplinary Journal of Nonlinear Science},
  30\penalty0 (10):\penalty0 103121, 2020.

\bibitem[Kennel et~al.(1992)Kennel, Brown, and Abarbanel]{Kennel92}
M.~B. Kennel, R.~Brown, and H.~D.~I. Abarbanel.
\newblock Determining embedding dimension for phase-space reconstruction using
  a geometrical construction.
\newblock \emph{Physical Review A}, 45:\penalty0 3403, 1992.

\bibitem[Baydin et~al.(2017)Baydin, Pearlmutter, Radul, and Siskind]{Baydin17}
A.~G. Baydin, B.~A. Pearlmutter, A.~A. Radul, and J.~M. Siskind.
\newblock Automatic differentiation in machine learning: a survey.
\newblock \emph{Journal of Machine Learning Research}, 18\penalty0
  (1):\penalty0 5595--5637, 2017.

\bibitem[Temam(2012)]{Temam12}
R.~Temam.
\newblock \emph{Infinite-Dimensional Dynamical Systems in Mechanics and
  Physics}.
\newblock {Springer Science \& Business Media}, 2012.

\bibitem[He et~al.(2016)He, Zhang, Ren, and Sun]{He16}
K.~He, X.~Zhang, S.~Ren, and J.~Sun.
\newblock Deep residual learning for image recognition.
\newblock In \emph{Proceedings of the IEEE conference on computer vision and
  pattern recognition}, 2016.

\bibitem[Ioffe and Szegedy(2015)]{Ioffe15}
S.~Ioffe and C.~Szegedy.
\newblock Batch normalization: Accelerating deep network training by reducing
  internal covariate shift.
\newblock In \emph{International conference on machine learning}, 2015.

\bibitem[Vincent et~al.(2008)Vincent, Larochelle, Bengio, and
  Manzagol]{Vincent2008a}
P.~Vincent, H.~Larochelle, Y.~Bengio, and P.-A. Manzagol.
\newblock Extracting and composing robust features with denoising autoencoders.
\newblock In \emph{Proceedings of the 25th international conference on Machine
  learning}, pages 1096--1103, 2008.

\bibitem[Vincent et~al.(2010)Vincent, Larochelle, Lajoie, Bengio, Manzagol, and
  Bottou]{Vincent10}
P.~Vincent, H.~Larochelle, I.~Lajoie, Y.~Bengio, P.-A. Manzagol, and L.~Bottou.
\newblock Stacked denoising autoencoders: Learning useful representations in a
  deep network with a local denoising criterion.
\newblock \emph{Journal of machine learning research}, 11\penalty0
  (12):\penalty0 3372--2408, 2010.

\bibitem[Srivastava et~al.(2014)Srivastava, Hinton, Krizhevsky, Sutskever, and
  Salakhutdinov]{Srivastava14}
N.~Srivastava, G.~Hinton, A.~Krizhevsky, I.~Sutskever, and R.~Salakhutdinov.
\newblock Dropout: a simple way to prevent neural networks from overfitting.
\newblock \emph{The journal of machine learning research}, 15\penalty0
  (1):\penalty0 1929--1958, 2014.

\bibitem[Kingma and Ba(2015)]{Kingma15}
D.~Kingma and J.~Ba.
\newblock Adam: A method for stochastic optimization.
\newblock In \emph{International Conference on Learning Representations}, 2015.

\bibitem[Brunton et~al.(2017)Brunton, Brunton, Proctor, Kaiser, and
  Kutz]{Brunton17}
S.~L. Brunton, B~W. Brunton, J.~L. Proctor, E.~Kaiser, and J.~N. Kutz.
\newblock Chaos as an intermittently forced linear system.
\newblock \emph{Nature Communications}, 8\penalty0 (19), 2017.

\bibitem[Wang et~al.(2021{\natexlab{a}})Wang, Yu, and Perdikaris]{Wang21a}
S.~Wang, X.~Yu, and P.~Perdikaris.
\newblock When and why pinns fail to train: A neural tangent kernel
  perspective.
\newblock arXiv:2007.14527, 2021{\natexlab{a}}.

\bibitem[Wang et~al.(2021{\natexlab{b}})Wang, Teng, and Perdikaris]{Wang21b}
S.~Wang, Y.~Teng, and P.~Perdikaris.
\newblock Understanding and mitigating gradient pathologies in physics-informed
  neural networks.
\newblock arXiv:2001.04536, 2021{\natexlab{b}}.

\end{thebibliography}

\end{document}